\documentclass[journal]{IEEEtran}

\usepackage{times}
\usepackage{epsfig}
\usepackage{graphicx}
\usepackage{amsmath}
\usepackage{amssymb}
\usepackage{subfigure}
\usepackage{epstopdf}
\usepackage{multirow}
\usepackage{longtable}
\usepackage{rotating}

\DeclareMathOperator*{\argmax}{argmax}

\ifCLASSINFOpdf
\else
\fi
\begin{document}
%
\title{A New Target-specific Object Proposal Generation Method for Visual Tracking}
%
%
%

\author{Guanjun~Guo,
        Hanzi~Wang*,~\IEEEmembership{Senior Member,~IEEE,}
        Yan~Yan,~\IEEEmembership{Member,~IEEE,} \\
        Hong-Yuan~Mark Liao,~\IEEEmembership{Fellow,~IEEE,}
        Bo~Li~\IEEEmembership{}
\IEEEcompsocitemizethanks{
\IEEEcompsocthanksitem *~Corresponding author. Tel./fax: +86 5922580063.\protect
\IEEEcompsocthanksitem G.~Guo, H.~Wang, and Y.~Yan are with the Fujian Key Laboratory of Sensing and Computing for Smart City, and the School of Information Science and Engineering, Xiamen University, Xiamen 361005, Fujian, P. R. China.\protect
~(E-mail: gjguo@stu.xmu.edu.cn; hanzi.wang@xmu.edu.cn; yanyan@xmu.edu.cn).
\IEEEcompsocthanksitem H. Y. Mark Liao is with the Institute of Information
Science, Academia Sinica, Taipei 115, Taiwan.\protect
 ~(E-mail: liao@iis.sinica.edu.tw).
 \IEEEcompsocthanksitem B. Li is with the Beijing Key Laboratory of Digital Media, and the School of Computer Science and Engineering, Beihang
University, Beijing 100191, P. R. China.\protect
 ~(E-mail: boli@buaa.edu.cn).
 }
\thanks{}}

%
%

\markboth{IEEE TRANSACTIONS ON CYBERNETICS}%
{Shell \MakeLowercase{\textit{et al.}}: Bare Demo of IEEEtran.cls for Journals}
%



\maketitle

\begin{abstract}
Object proposal generation methods have been widely applied to many computer vision tasks. However, existing object proposal generation methods often suffer from the problems of motion blur, low contrast, deformation, etc., when they are applied to video related tasks. In this paper, we propose an effective and highly accurate target-specific object proposal generation (TOPG) method, which takes full advantage of the context information of a video to alleviate these problems. Specifically, we propose to generate target-specific object proposals by integrating the  information of two important objectness cues: colors and edges, which are complementary to each other for different challenging environments in the process of generating object proposals. As a result, the recall of the proposed TOPG method is significantly increased. Furthermore, we propose an object proposal ranking strategy to increase the rank accuracy of the generated object proposals. The proposed TOPG method has yielded significant recall gain (about 20\%-60\% higher) compared with several state-of-the-art object proposal methods on several challenging visual tracking datasets. Then, we apply the proposed TOPG method to the task of visual tracking and propose a TOPG-based tracker (called as TOPGT), where TOPG is used as a sample selection strategy to select a small number of high-quality target candidates from the generated object proposals. Since the object proposals generated by the proposed TOPG cover many hard negative samples and positive samples, these object proposals can not only be used for training an effective classifier, but also be used as target candidates for visual tracking.  Experimental results show the superior performance of TOPGT for visual tracking compared with several other state-of-the-art visual trackers (about 3\%-11\% higher than the winner of the VOT2015 challenge in term of distance precision).
\end{abstract}


%
\IEEEpeerreviewmaketitle

\section{Introduction}
\IEEEPARstart{O}{bject} proposal generation methods aim to generate a small number of object proposals (i.e., bounding boxes) to guide the search for objects in images. Since the computational complexity of locating an object can be significantly reduced by focusing only on the generated object proposals (instead of the bounding boxes with all possible positions and scales obtained by using the sliding window strategy), object proposal generation methods have been widely applied to object detection~\cite{Rcnn2014,XiaozhiNIPS15}, visual tracking~\cite{hua2015online,kcfdp2015,ZhuPL16} and other computer vision tasks~\cite{Meng_2016_CVPR,Johnson_2016}. Usually, object proposals are generated by using either the color cue or the edge cue. Representative methods include Selective Search~\cite{SelectiveSearch13} and EdgeBoxes~\cite{EdgeBoxes2014}. However, it is difficult to achieve high recall when only one cue is used. For example, the color based object proposal generation methods may be less effective when they are applied to gray images. And the edge based object proposal generation methods may not work well when objects have weak edges. Recently, several deep learning based object proposal generation methods~\cite{KongCVPR16,ren2015faster} have been developed. However, these methods usually have a large computational burden. In addition, existing object proposal generation methods may fail to obtain target regions when they are applied to the task of visual tracking, since the targets in a video often suffer from the problems of motion blur, low contrast, occlusion, etc. Moreover, most of existing object proposal generation methods are usually designed to generate object proposals for complete objects instead of parts of objects, which limits their applications to visual tracking.
\begin{figure}
   \begin{center}
   \includegraphics[width=0.48\textwidth]{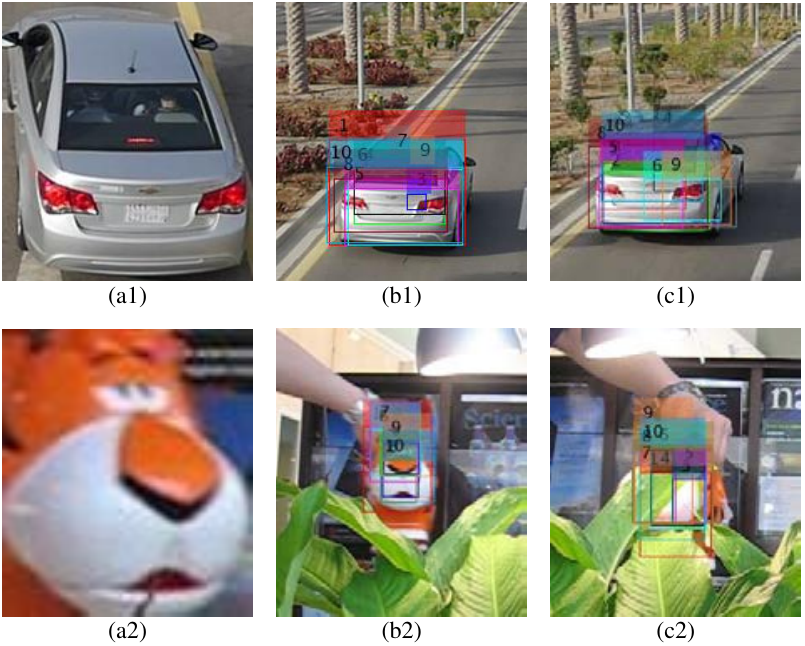}
   \end{center}
   \vspace{-0.251cm}
   \caption{\label{overviewFig}Illustration of the proposed target-specific object proposal generation method to generate object proposals on two different test images (in the first and second rows, respectively). (a1)/(a2) The specified targets. (b1/b2) and (c1/c2) The top ten object proposals obtained by the proposed TOPG method on two test images for each specified target.}
\end{figure}
In this paper, we propose a novel target-specific object proposal generation (TOPG) method, which is specially designed for the task of visual tracking.  The proposed method uses multiple cues to generate object proposals for a given target, by which the recall is significantly increased compared with several existing object proposal
generation methods.  We also propose a ranking strategy to rank the generated object proposals by using multiple prior information of a specified target.  The proposed ranking strategy can effectively suppress noisy object proposals.  As a result, when a fixed number of ranked object proposals are used for evaluation, the accuracy can be significantly increased while the recall can be satisfactorily retained.  In other words, the proposed method can effectively recall correct instances even when these target instances suffer from bad conditions like motion blur, deformation, low contrast and occlusions.

In this paper, two commonly used objectness cues, i.e. color and edge, are jointly utilized to generate target-specific object proposals in the proposed TOPG method.  Probability distributions of the pixels in a target region and those in the background are respectively estimated using a non-parametric density estimation method.  We then estimate the probability of each pixel belonging to the target region, by which the response map of a test image is generated.  Since the response map is generated based on the color information, it is less affected by weak edges.  When we make use of the edges detected from the original image as well as the above-mentioned response map to generate object proposals, the target recall can be significantly increased compared to the approaches that use the edges extracted from the original image.  At last, we propose to use the combined affinity of shape, color and size between a generated object proposal and the given target to rank all the generated object proposals.  Fig.~\ref{overviewFig} shows a number of best-ranked object proposals selected by the proposed TOPG method.  Fig.~\ref{overviewFig} (a1) and (a2) respectively show two specified targets.  The top ten ranked object proposals corresponding to the two given targets on two different test images are respectively shown in (b1)/(c1) and (b2)/(c2) in Fig.~\ref{overviewFig}. As can be seen, the targets can be effectively recalled by using the proposed TOPG method.

Since the object proposals generated by the proposed TOPG can effectively cover the target specified by a user, we propose a TOPG-based tracker (call as TOPGT) following the tracking-by-detection framework.  Unlike the conventional trackers~\cite{hua2015online,kcfdp2015,ZhuPL16} that adopt the Edgebox algorithm, our TOPGT directly uses the object proposals generated by the TOPG method.  The new TOPGT tracker can effectively boost the tracking performance since the TOPG method can obtain more accurate object proposals to train its classifier and thus achieve higher recall than the conventional Edgebox-based trackers.  Besides, a classifier trained by the object proposals generated by TOPG has strong discriminative power for tracking since these proposals usually suggest similar objects, including hard positive samples (e.g., deformed samples and occluded samples) and hard negative samples (e.g., similar objects in background).  Another superior capability of the proposed TOPGT is its robustness for long-term tracking. This is because the object proposals adopted by TOPGT are inherently robust for scale variations, occlusions, and appearance changes.

The main contributions of this work are summarized as follows: We propose a novel target-specific object proposal generation (TOPG) method.  In comparison with existing state-of-the-art object proposal generation methods, our TOPG method can achieve the best recall rate.  In addition, the TOPG method can generate high-quality object proposals since these proposals contain better hard negative samples  for training.  Thus, we can use these samples to train a more robust classifier for visual tracking.  We also theoretically analyze the superiority of the proposed TOPG method over several object proposal generation methods for visual tracking.  Extensive experiments conducted on two challenging visual tracking datasets (i.e., OTB-2013~\cite{cvpr2013} and UAV20L~\cite{uav123}) demonstrate that the proposed TOPGT outperforms several state-of-the-art trackers.

The remainder of this paper is organized as follows: Section II gives an overview of several existing object proposal generation methods and several related trackers. Section III represents the details of the proposed TOPG method. Moreover, the details of how the proposed TOPG method is applied to the task of visual tracking are also given. Section IV describes the experimental results obtained by TOPG for object proposal generation and the  experimental results obtained by the proposed TOPG-based tracker on several challenging visual tracking datasets.  Section V draws the conclusions.

\section{Related Work}
In this section, we first discuss several representative object proposal generation methods. Then, we discuss several related visual trackers.

\textbf{Object Proposal Generation Methods.}
Existing object proposal generation methods can be roughly divided into two categories according to the way of generating object proposals. The first category of the object proposal generation methods are based on image segmentation, and these methods treat different image segments as object proposals. For example, Selective Search~\cite{SelectiveSearch13} generates image segments by using a superpixel segmentation algorithm in different color spaces, and then greedily merges the segments to generate object proposals.
MCG~\cite{MCG2014} introduces a multi-scale hierarchical segmentation algorithm to segment images  based on edges, and then merges the generated segments based on the edge strength. Usually, the object proposals are finally ranked by using multiple objectness cues, such as size, location, shape and edge strength. Many image segmentation based object proposal generation methods (such as Geodesic~\cite{geodesic2014}, CPMC~\cite{CPMC2012}, Chang~\cite{Chang2011} and CM~\cite{guo2017}) follow this scheme. However, due to the highly computational complexity of the employed image segmentation algorithms, these segmentation based object proposal generation methods are usually time-consuming.

The second category of the object proposal generation methods are  to assign different scores to different object proposals. These methods sample a large number of candidate windows and utilize the scoring functions to give each candidate window a score. These methods usually have a faster running speed compared with the first category of object proposal generation methods. EdgeBoxes~\cite{EdgeBoxes2014}, Cascaded SVM~\cite{twoSVM}, FCN~\cite{jie2016}, OOP~\cite{OOP2015} and BING~\cite{BING2014} are several representative methods of this category. The EdgeBoxes algorithm is closely related to the proposed TOPG method. The scoring function of the EdgeBoxes algorithm is heuristically formulated based on the number of edges contained in a candidate window. Thus, EdgeBoxes usually gives a high score to a candidate window, which contains a large number of edges. However, EdgeBoxes may fail to locate objects that have weak edges. The proposed TOPG method can overcome this drawback by using complementary cues (i.e., color/intensity and edge cues). The response map corresponding to a test image is firstly obtained by estimating the probability of each pixel in the test image belonging to a tracked target using the color/intensity cue. The target in the resulting response map usually has strong edges even when the target suffers from the problems of motion blur, low contrast and so on. Then, object proposals are generated by TOPG on both the response map and the test image by using the edge cue to recall the target.
Moreover, the Cascaded SVM, FCN and BING methods usually generate object proposals for a complete object instead of a part of the object, which limits their applications to the task of visual tracking. In contrast, the proposed TOPG method can be used to generate object proposals for an arbitrary part of a target region.

\begin{figure*}
\begin{center}
   \includegraphics[width=0.95\linewidth]{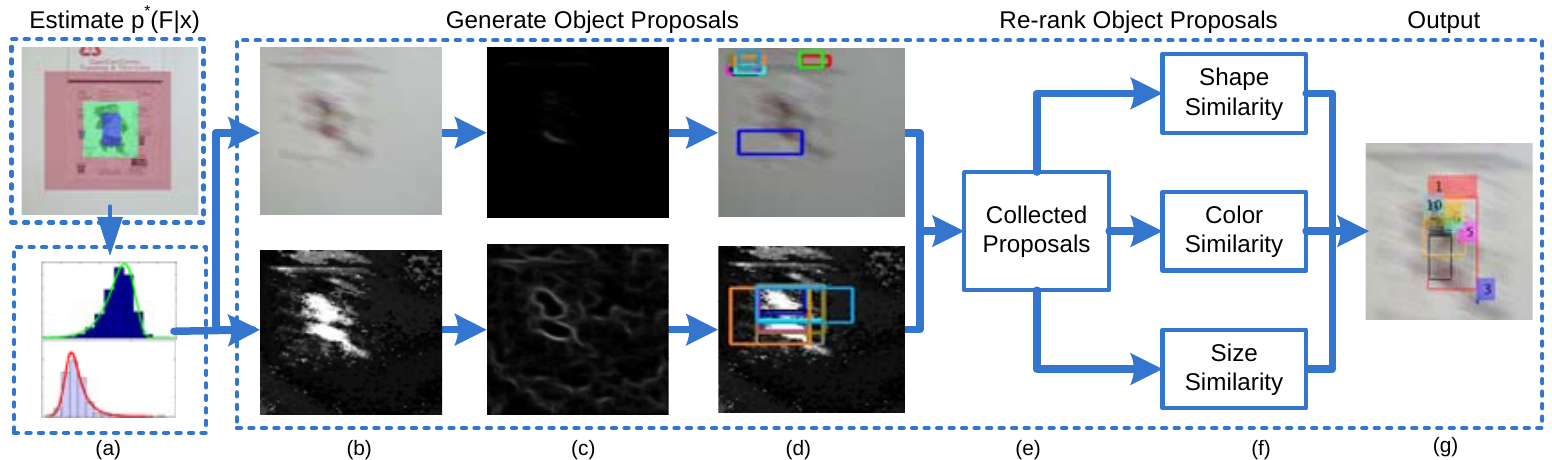}
\end{center}
   \caption{The framework of the proposed TOPG method. (a) Estimating the probability distributions (shown in the second row) of the pixels from the foreground and those from the background by using the tri-map (shown in the first row) of an input image, respectively. (b) The test image (top) and its response map (bottom). (c) The edges obtained from the test image and the response map, respectively. (d) The generated object proposals based on the obtained edges. (e)-(f) Ranking the object proposals by using the proposed object proposal ranking strategy. (g) The top-ten ranked object proposals obtained by the proposed TOPG method.}
\label{Fig::intuition}
\end{figure*}
\textbf{Visual Tracking Methods.}
Tracking-by-detection methods have shown good performance in recent years. The goal of the tracking-by-detection methods is to use an effective classifier, which is trained by using online learning techniques, to detect a tracked target. Representative works are TLD~\cite{TLD2012}, Struck~\cite{Struck2011}, KCF~\cite{Henriques2015}, MDNet~\cite{MDNet2016}, GM~\cite{lu2016}, etc. However, it is tricky to select samples in the process of training a classifier in the tracking-by-detection framework, since the trade-off between computational cost and good performance needs to be considered. Recently,
correlation filter based tracking methods (such as~\cite{bolme2010,Henriques2015,martin14,yangli2015,martin2016,sui2017,kristan2017}) have shown promising performance for visual tracking. Usually, these methods train a bank of filters as a classifier from all the shifts of a target within only a few milliseconds, since the shifts of the target can be represented by one base target in the Fourier domain using the properties of the circulant matrices~\cite{Henriques2012}.

Moreover, due to the success of CNN on many computer vision tasks, CNN features have been applied to correlation filter based methods (such as C-COT~\cite{CCOT},  ECO~\cite{ECO17}, FCNT~\cite{FCNTracker2015}, HCF~\cite{ma2015}, TPM~\cite{zhangyi2017}, etc.). These methods have been demonstrated effective for visual tracking. In addition, several visual tracking methods based on CNN have been proposed, such as~\cite{zhangyi2017,MDNet2016}. For example,
MDNet~\cite{MDNet2016} proposes a multi-domain learning strategy to train a CNN classifier based on a set of image sequences, where the trained CNN classifier can effectively discriminate the tracked target from the background. However, MDNet may easily drift when similar objects appear in neighboring regions, since the CNN classifier used in MDNet is trained by using the samples drawn from the specified Gaussian distribution without considering of other training samples, such as distractors and similar objects. SANet~\cite{SANet17} improves the discriminative ability of the CNN classifier used in MDNet. SANet is able to handle similar distractors by modeling the self-structure of an object with a recurrent neural network. However, both MDNet and SANet suffer from the problem that similar distractors are hard to be sampled during training the CNN classifier. In contrast, the proposed TOPG method can be treated as an effective sample selection strategy for the task of visual tracking. More specifically, the proposed TOPG method can be seamlessly integrated into the above trackers, where the problem of sample selection is formulated as the problem of target-specific object proposal generation, and the tracking problem is treated as the task of selecting the best object proposal from the object proposals generated by TOPG.

\section{Target-specific Object Proposal Generation}
 In this section, we first provide the details of the proposed TOPG method in Sec. III-A. Then, we provide the details of the proposed object proposal ranking strategy in Sec. III-B. In Sec. III-C, we show how the proposed TOPG method is applied to the task of visual tracking.

\subsection{Target-specific Object Proposal Generation via Multiple Cues}
\label{sec:TOPG}
Existing object proposal generation methods use either edge or contour information to generate object proposals since these features have the virtue of high computational efficiency.  However, video frames usually contain weak edges due to motion blur, low contrast, cluttered background, etc.  Conventional object proposal generation methods may not be able to generate appropriate object proposals for objects with weak edges.  Therefore, we propose to use two complementary cues, i.e. color (or intensity for gray images) and edge~\cite{Objectness2012,guo2017}, to generate object proposals.  The detailed procedure of the proposed TOPG method is illustrated in  Fig.~\ref{Fig::intuition}.  Given a bounding box in the first row of Fig.~\ref{Fig::intuition}(a), which contains a target region (in the blue color) and the background region (in the pink color), the probability distributions of the target and the background are respectively estimated and shown in the second row of Fig.~\ref{Fig::intuition}(a).  Then a response map (shown in the second row of Fig.~\ref{Fig::intuition}(b)) is generated by calculating the probability of each pixel in the target region. The response map is complementary to the original image for generating object proposals by using the edge information, since the resulting response map is produced by using the color information and it can highlight the target specified by a user.  Fig.~\ref{Fig::intuition}(c) shows the edges detected from the original bounding box (top) and its corresponding response map (bottom), respectively.  It shows that the edges detected from the response map are stronger and they can better highlight the target.  We then propose to collect object proposals based on the edges generated in the original bounding box and the response map (which are respectively shown in Fig.~\ref{Fig::intuition}(d)).  Fig.~\ref{Fig::intuition}(e)-(g) illustrate how these proposals are ranked.  The whole procedure is elaborated as follows.

In order to avoid mislabeling the pixels from a target and the surrounding background when computing their probability distributions, each training image is partitioned into three regions: a definite foreground, a definite background and a blended region (where the pixels from the target and the background are mixed).
As shown by the green region in the first row of Fig.~\ref{Fig::intuition}(a), the blended region is a hollow rectangle (in the green color) whose height/width is $\gamma$ ($1<\gamma<2$) times the height/width of the target, where a part of the blended region is from the target and the  remaining part is from the background. The definite foreground and the definite background are the region inside (in the blue color) and the region outside (in the pink color) of the hollow rectangle (i.e., the blended region), respectively.
Let $x^f$ and $x^b$ respectively denote the pixels from the definite foreground (i.e., the target) $F$ and the pixels from the definite background $B$. Given $x^f$ and $x^b$, the probability distribution of $x^f$ (denoted by $p^f$) and the probability distribution of $x^b$ (denoted by $p^b$) are respectively estimated by using the histogram of $x^f$ and that of $x^b$. Given a test image $x$, the probability of each pixel of the test image belonging to the foreground can be approximated as follows.
\begin{equation}\label{eq::foreProbability}
  p^*(F|x)=\frac{p^f(x)}{p^f(x)+p^b(x)}.
\end{equation}
Since the appearances of a target and the background may change in the task of visual tracking, the probability distributions $p^f(x)$ and $p^b(x)$ (written as $p^{f/b}(x)$ for simplicity) are updated every $\kappa$ frames. It can be written as
  $p_m^{f/b}(x)=\lambda  p_m^{f/b}(x)+(1-\lambda)p_{m-\kappa}^{f/b}(x)$,
where $\lambda$ is a learning rate, and $m$ denotes the index of the current video frame. Such an updating scheme can be used to correctly estimate the probability distributions when the appearances of the target and the background change.

\begin{figure*}
   \begin{center}
  \includegraphics[width=0.90\linewidth]{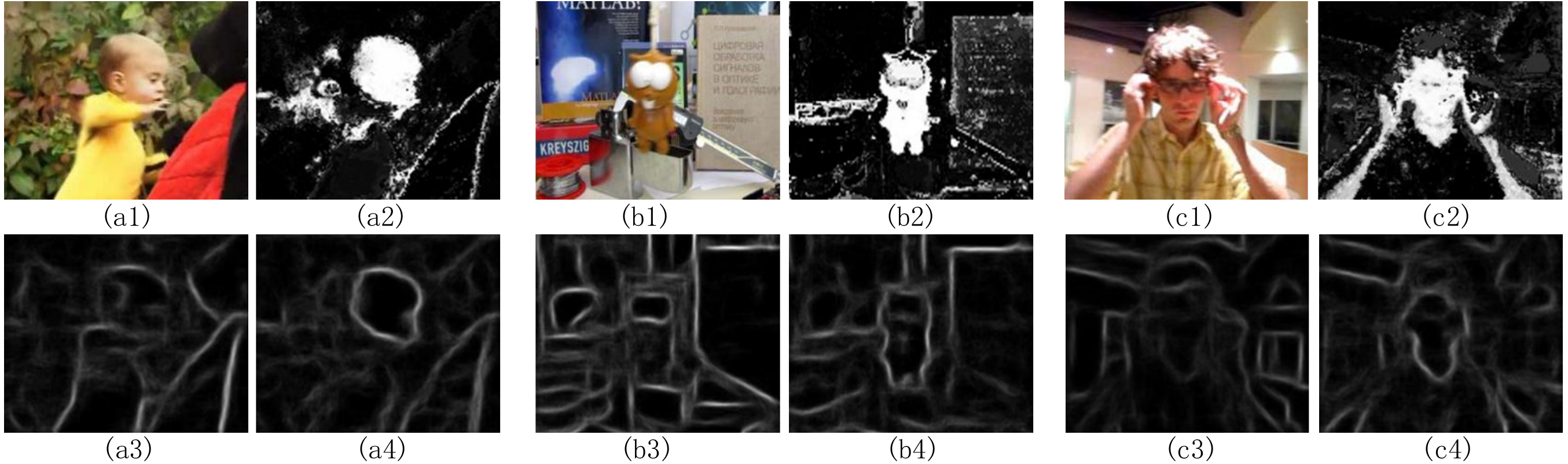}
   \end{center}
   \vspace{-0.15cm}
   \caption{\label{Fig::edgeCom}The comparison of the edge images obtained from the original images and their response maps. (a1)/(b1)/(c1) The three input images. (a2)/(b2)/(c2) The corresponding  response maps. (a3)/(b3)/(c3) and (a4)/(b4)/(c4) The edge images detected on the input images and the response maps, respectively.}
\end{figure*}
The response map is produced by using Eq.~(\ref{eq::foreProbability}) on each pixel in the test image. Although the response map may contain several false predictions, it can efficiently suppress most of the noisy information from the background and highlights the target (see Fig.~\ref{Fig::edgeCom}).  Thus, the response map can be used to generate high-quality object proposals. Inspired by the EdgeBoxes~\cite{EdgeBoxes2014} and gPb-owt-ucm~\cite{gPbowtucm2011} methods, we use the information of edges to quickly generate object proposals. The structured edge detector \cite{edgeDet2015}, due to its effectiveness and efficiency, is employed to detect edges. To obtain all possible object proposals for the target, the original image is also used to detect edges. The response map and the original image are complementary to each other for detecting strong edges.  The edges in the target region obtained  from the original image may contain uninformative weak edges when the target suffers from the problems of motion blur, low contrast, background clutter, and so on. However, the edges obtained from the response map usually contain the strong edges of the target. Fig.~\ref{Fig::edgeCom} shows a qualitative comparison of the edge images detected from three original images (shown in Fig.~\ref{Fig::edgeCom}(a1)/(b1)/(c1), respectively) and their response maps (shown in Fig.~\ref{Fig::edgeCom}(a2)/(b2)/(c2), respectively). The edge images detected on the corresponding  input images and response maps are shown in the second row of Fig.~\ref{Fig::edgeCom}. As can be seen, the edge images detected from the response maps effectively highlight the targets.


EdgeBoxes is adopted in this paper to generate object proposals. EdgeBoxes generates object proposals based on the observation that the number of contours enclosed
by a bounding box is a likelihood of the bounding box containing an object. In the implementation of EdgeBoxes, a value (denoted as $\mu \in [0,1]$) that indicates whether an edge is contained in a bounding box, is calculated. Then the score of an object proposal $\rho$ is defined by using the normalized sum of all the $\mu$ values in the bounding box. The score $\rho$, which measures how likely a bounding box contains an object, is used in the proposed object proposal ranking strategy as the shape affinity between an object proposal and a specified target.

To generate target-specific object proposals, we integrate the EdgeBoxes algorithm, which is only used to generate object proposals, into the proposed TOPG method.  Other scoring-based object proposal generation methods can also be integrated into our method.  However, the recall achieved by our method is higher than that achieved by existing scoring-based methods.  A further analysis about why our method achieves higher recall than other scoring-based methods is elaborated in Appendix A.

\begin{table*}\small
\caption {\label{tbCNNStru}The structure of the CNN net used in the proposed tracker.}
\begin{center}
\begin{tabular}{  l | l  l  l l l l l l l }
  \hline
  Layer   & Layer1      & Layer2 & Layer3        & Layer4  & Layer5       &Layer6  &Layer7 &Layer8  &Layer9      \\ \hline
  Net1    & C(7, 2, 3, 96) & P(2, 2)   & C(5, 2, 96, 256)  & P(2, 3)    & C(3, 1, 256, 512) &C(3, 1, 512, 512) &F(512) &F(512) &F(2) \\
  \hline
\end{tabular}
\end{center}
\end{table*}
\subsection{Target-specific Proposal Ranking Strategy}
As discussed in Section~\ref{sec:TOPG}, high target recall can be achieved by generating object proposals on the search area containing a target in both the test image and its corresponding response map.  However, it is possible to generate many redundant object proposals.  Therefore, we propose an object proposal ranking strategy to rank all generated object proposals, by which high recall can be retained when a small number of top-ranked object proposals are used for evaluation.  The proposed ranking strategy ranks all generated object proposals based on a set of affinity values, which are calculated between each object proposal and the target region using three cues, including shape, color, and size.

\textbf{Shape affinity.} Since  the likelihood of the number of contours enclosed by a bounding box $\rho$ (used in EdgeBoxes)  can be treated as a shape score, the shape affinity value $s_{i,t}$ between the $i$th object proposal and a target region $\tau_t$ is defined as:
 \begin{equation}\label{eq::shapeAffinity}
  s_{i,t}=e^{(-|\rho_i-\rho_t|)},
\end{equation}
where $\rho_i$ and $\rho_t$ respectively denote the shape score of the $i$th proposal and the target.

\textbf{Color affinity.} The color affinity $c_{i,t}$ between the $i$th object proposal and the target can be computed by using a response map. The mean of all the values of the pixels from the region of the response map, which corresponds to the $i$th object proposal, is computed as the color affinity value between the $i$th object proposal and the target.

\textbf{Size affinity.} The target size is also key prior information and it can be used to filter out oversized or undersized object proposals. Similar to the definition of the shape affinity, the size affinity value $z_{i,t}$ between the $i$th proposal and the target is defined as:
 \begin{equation}\label{eq::shapeAffinity}
  z_{i,t}=e^{(-|w_i-w_t|)} \cdot e^{(-|h_i-h_t|)},
\end{equation}
where $w_i$/$h_i$ and $w_t$/$h_t$ respectively denote the width/height of the $i$th object proposal and the target.

Since the above three affinities are independent to each other, the combined affinity value $a_{i,t}$ between the $i$th proposal and the target can be defined as the product of the three affinities, which is given by
 \begin{equation}\label{eq::Affinity}
  a_{i,t}= s_{i,t}\cdot c_{i,t}\cdot z_{i,t}.
\end{equation}
The generated object proposals are ranked according to the combined affinity values in descending order. The proposed ranking strategy can effectively filter out noisy object proposals. As a result,  only a small number of high-quality ranked object proposals are retained for maintaining high recall.

\subsection{Application of TOPG to Visual Tracking}
Since the proposed TOPG method can effectively recall a specified target, it may be applied to most existing visual object trackers as the sample selection component. In this subsection, we propose a TOPG-based tracker (i.e., TOPGT) for object tracking.
The proposed TOPGT employs the framework of tracking-by-detection.  However, different from the CNN classifier used in other trackers (such as MDNet~\cite{MDNet2016}), where the CNN classifier is trained by drawing training samples from a specified probability distribution, the objects proposals generated by TOPG on the previous frame are used to train the CNN classifier in TOPGT.  The generated object proposals usually contain some candidate windows similar to the target (i.e., hard negative samples). Therefore, the classifier trained from these object proposals is not sensitive to the background distractors, which are similar to the tracked target.

The CNN structure adopted in TOPGT is shown in Table~\ref{tbCNNStru}. In Table~\ref{tbCNNStru}, the convolutional layer is denoted by C($q$, $d$, $nIn$, $nOut$), where  the kernel size of the filters is $q\times q$, $d$ denotes the stride, $nIn$ and $nOut$ respectively denote the number of the input feature maps and the number of the output feature maps. Each convolutional layer is activated by a rectified linear unit (ReLU)~\cite{ReLU2013}. P($q$, $d$) denotes the max-pooling layer, whose kernel size is $q\times q$. F($o$) denotes the fully connected layer and the number of the output nodes is $o$. The first six layers of the CNN structure in Table~\ref{tbCNNStru} are the same as the corresponding layers of the VGG-M net~\cite{Chatfield14}, while the last three layers of the CNN structure of TOPGT are the fully connected layers, which is used as a classifier.

\begin{figure*}
\begin{center}
  \includegraphics[width=0.90\linewidth]{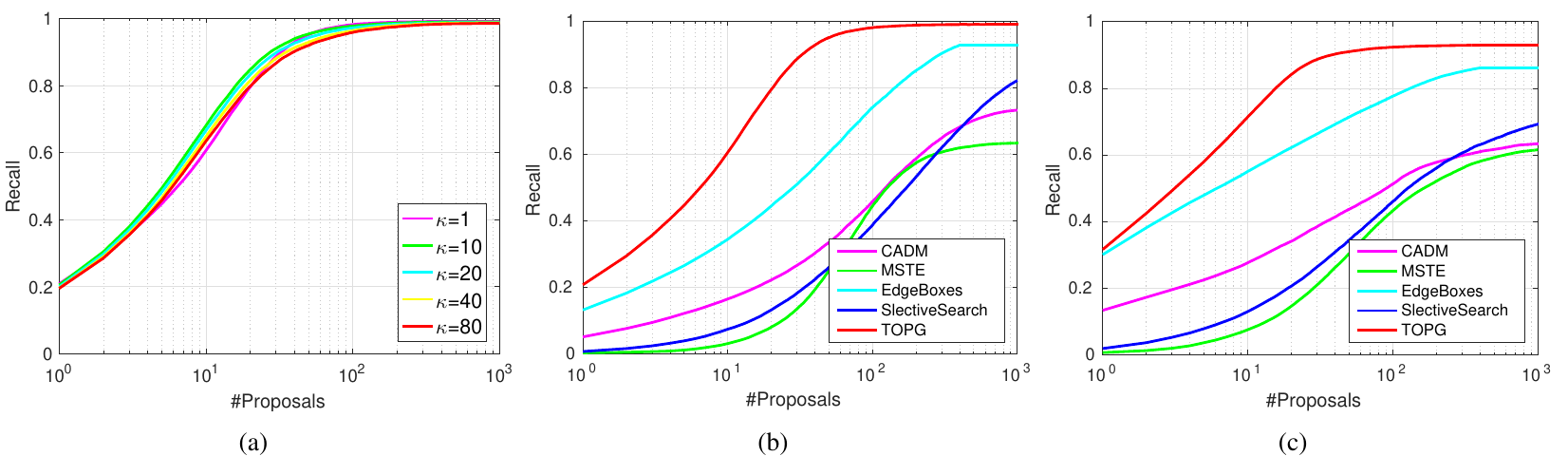}
\end{center}
\vspace{-0.251cm}
   \caption{\label{fig:ComparingRecallCurves} The recall curves obtained by the proposed TOPG method and the other four competing object proposal generation methods. (a) The trade-off between the number of object proposals in each image and the recall obtained by the proposed TOPG method for object proposal generation when varying the value of the interval $\kappa$ on the OTB-2013 dataset. (b)-(c) The trade-off between the number of object proposals in each image and the recall for object proposal generation on the OTB-2013 dataset and the UAV20L dataset, respectively.}
\end{figure*}
In the procedure of training the CNN classifier in TOPGT, we firstly initialize the weights of all the four convolutional layers shown in Table~\ref{tbCNNStru} by using the corresponding weights obtained by the VGG-M net, which is trained  on the ILSVRC-2012 dataset~\cite{ILSVRC15}.  The weights of the three fully connected layers are initialized with a Gaussian distribution with a zero mean and the variance equal to $10^{-2}$. Effective CNN features can be extracted by using the corresponding weights from the VGG-M net. Then, we fine-tune the CNN classifier by using the object proposals generated by TOPG, by which the CNN classifier can  discriminate a target from the background. If the CNN classifier in the proposed tracker is not fine-tuned by using the object proposals generated by TOPG, the tracker may be easily distracted by the objects that are similar to the tracked target. The main reason is that the VGG-M net is trained for the task of image classification, which mainly focuses on inter-class distinguishability instead of target distinguishability. In our implementation, both positive samples and negative samples are selected from the object proposals generated by TOPG by using a threshold on the Intersection over Union (IoU) ratio. We define the IoU ratio as $\frac{PBB \bigcap TBB}{PBB \bigcup TBB}$,
where $PBB$ and $TBB$ denote the bounding box of a generated proposal window and the target region derived from the previous frame, respectively. The positive samples and the negative samples are the object proposals having IoU $>\varphi$ or IoU $<\omega$, respectively.
Because that the number of positive samples may be far less than that of negative samples, which will cause the class imbalance problem, we also sample $\tilde{N}$ target candidates around the location of the tracked target in the previous frame for data augmentation. The $\tilde{N}$ target candidates are  generated by adding Gaussian noise to the bounding box of the tracked target in the previous frame (i.e., [$\tilde{x}$, $\tilde{y}$, $\tilde{w}$, $\tilde{h}$]), where $\tilde{x}$, $\tilde{y}$ respectively denote the horizontal and vertical coordinates of the top-left corner, and $\tilde{w}$, $\tilde{h}$ respectively denote the width and height of the target bounding box.   The mean of the Gaussian noise is set to zero and the covariance matrix is diag($0.01\tilde{w}^2$,$0.01\tilde{h}^2$,$0.01\tilde{w}^2$,$0.01\tilde{w}^2$).

In the tracking procedure, the object proposals, which are generated by TOPG from the search window around the target tracked in the previous frame, are used as the target candidates. The trained CNN classifier is then used to predict the probability of each object proposal being the tracked target. The tracked  target is located by choosing the object proposal, which has the highest probability. Since the object proposals generated by TOPG are robust to occlusions, scale variations, serious deformation, etc., the proposed TOPGT can obtain excellent tracking performance even in challenging scenarios.

\section{Experiments}
In this section, we firstly evaluate the performance of the proposed TOPG method for object proposal generation, and compare it with several state-of-the-art object proposal generation methods. Then we demonstrate the performance of the proposed TOPG-based tracker and compare it against several state-of-the-art visual trackers. We use two recently published challenging visual tracking datasets: OTB-2013~\cite{cvpr2013} and UAV20L~\cite{uav123}, to conduct both the object proposal generation and  visual tracking experiments. The OTB-2013 dataset consists of 50 test video sequences with different annotated attributes, which include  \emph{background clutter}, \emph{occlusions}, \emph{motion blur}, etc. The UAV20L dataset consists of 20 annotated high-definition videos captured by a drone from a low-altitude aerial perspective, and it is specially collected for the task of long-term visual tracking. In the UAV20L dataset, the shortest and longest video sequences have $1,717$ and $5,527$ frames, respectively.

For the task of object proposal generation, we adopt the Intersection over Union (IoU) metric~\cite{cvpr2013} as the evaluation measure. IoU is set to 0.5, which is used as a threshold to check whether an object proposal hits a target. The number of the bins of the histograms, which are used to estimate the probability distributions of the pixels from the target and the background, is set to 32. The number of proposal windows generated by EdgeBoxes is set to $1000$. $\gamma$ is set to 0.4. $\tilde{N}$, which denotes the number of target candidate windows to be sampled from a Gaussian distribution, is set as $100$.

For the task of visual tracking, we report the distance precision (DP) and the success rate results in one-pass evaluation (OPE) and spatial robustness evaluation (SRE) criteria introduced as in~\cite{cvpr2013}.
DP is the relative number of frames in a video sequence, where the Euclidean distance between the center location of the estimated target and the ground truth center location of the target is smaller than a threshold.
The success rate is defined as the percentage of frames, where the overlap between the estimated bounding box and the ground truth bounding box is greater than a threshold $\eta\in (0,1]$.
Following~\cite{martin2016,Henriques2015,cvpr2013}. the DP values are reported at a threshold of 20 pixels, and the success rates are reported by setting the overlap threshold $\eta$ to $0.5$. SRE tests the robustness of a tracker with different initial positions by slightly shifting or scaling the ground truth bounding boxes.

\begin{figure*}
\begin{center}
   \includegraphics[width=0.98\linewidth]{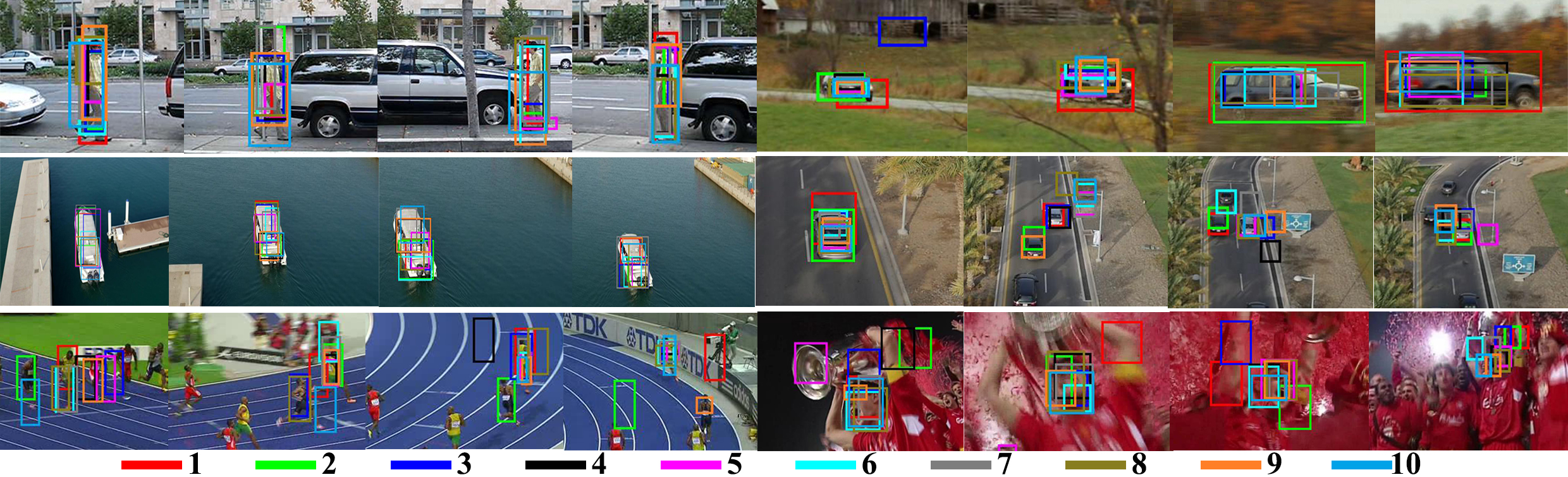}
\end{center}
   \vspace{-0.351cm}
   \caption{Qualitative target-specific object proposal generation results obtained by the proposed TOPG method on some videos of the OTB-2013 and the UAV20L datasets. Only top ten generated proposals are shown in the test images.}
\label{Fig:proposalDe}
\end{figure*}


\subsection{Evaluation of TOPG for Object Proposal Generation}
First, we evaluate how the value of the interval $\kappa$, which is used to update the foreground probability distributions (see Eq.~(\ref{eq::foreProbability})), affects the recall obtained by the proposed TOPG method when varying the number of object proposals. We empirically set the weight factor $\lambda$ to be a constant value 0.01. Fig.~\ref{fig:ComparingRecallCurves}(a) shows the results obtained by the TOPG method when using different $\kappa$ values. As can be seen, the recall obtained by the proposed TOPG method decreases slightly as  the value of $\kappa$ increases when the number of object proposals is greater than 100. The reason is that the foreground probability distributions are less frequently updated when the value of $\kappa$ is high. Nevertheless, the proposed TOPG method still performs well, which means it is robust to appearance variations.
In the following experiments, the value of $\kappa$ is empirically set to $30$ to balance the recall and the computational complexity.

\begin{table}
\caption{\label{tab:otb2013}The recall obtained by the proposed TOPG method and the other four competing methods for target-specific object proposal generation when varying the number of object proposals on the OTB-2013 dataset.}
\small
\begin{center}
\begin{tabular}{| c | c | c | c |c|c|}
    \hline
    \multirow{2}{*}{Methods} & \multicolumn{5}{c|}{ Number of Object Proposals} \\
    \cline{2-6}
                     & 50     & 100    & 200 & 500   &1,000 \\ \hline \hline
    CADM             & 0.335 & 0.458 & 0.587 & 0.703 & 0.733\\ \hline
    MSTE             & 0.250 & 0.444 & 0.574 & 0.626 & 0.634\\ \hline
    EdgeBoxes        & 0.605 & 0.741 & 0.852 & 0.928 & 0.928\\ \hline
    SelectiveSearch & 0.257 & 0.383 & 0.528 & 0.726 & 0.834\\ \hline
    TOPG            & \textbf{0.951} & \textbf{0.981} & \textbf{0.989} & \textbf{0.991} & \textbf{0.991}\\
    \hline
\end{tabular}
\end{center}
\end{table}

\begin{table}
\caption{\label{tab:uav20l}The recall obtained by the proposed TOPG method and the other four competing methods for target-specific object proposal generation when varying the number of object proposals on the UAV20L dataset.}
\small
\begin{center}
\begin{tabular}{| c | c | c | c |c|c|}
    \hline
    \multirow{2}{*}{Methods} & \multicolumn{5}{c|}{ Number of Object Proposals} \\
    \cline{2-6}
                     & 50     & 100    & 200 & 500   &1,000 \\ \hline \hline
    CADM             & 0.436 & 0.512 & 0.577 & 0.617 & 0.633\\ \hline
    MSTE             & 0.304 & 0.432 & 0.521 & 0.592 & 0.615\\ \hline
    EdgeBoxes        & 0.714 & 0.776 & 0.830 & 0.861 & 0.861\\ \hline
    SelectiveSearch & 0.326 & 0.447 & 0.556 & 0.652 & 0.702\\ \hline
    TOPG            & \textbf{0.910} & \textbf{0.923} & \textbf{0.927} & \textbf{0.929} & \textbf{0.929}\\
    \hline
\end{tabular}
\end{center}
\end{table}
We compare the proposed TOPG method with several state-of-the-art object proposal generation methods in terms of recall. We select the following four methods for comparison: CADM~\cite{CVPR15CADM}, MSTE~\cite{cvpr15mtse}, EdgeBoxes~\cite{EdgeBoxes2014} and Selective Search~\cite{SelectiveSearch13}. The four competing methods are selected since they are closely related to the proposed TOPG method: Selective Search and CADM are based on image segmentation; MSTE and EdgeBoxes are based on the boundary or the edge feature.
Fig.~\ref{fig:ComparingRecallCurves}(b) shows the comparison results on the OTB-2013 dataset. As can be seen from Fig.~\ref{fig:ComparingRecallCurves}(b) and Table~\ref{tab:otb2013}, the recall obtained by the proposed TOPG method is the highest among those obtained by the competing methods, and it is higher than that of the other competing methods by 34\%-70\% when the top 50 ranked object proposals are used for calculating the recall.
Although we only use the top 50 ranked object proposals for calculating the recall, the results we obtain are still better than those obtained by the competing methods even when they use the top 1000 ranked object proposals (about 2\% - 31\%).

The proposed TOPG method achieves excellent performance since it uses two cues (i.e., color and edge) to generate object proposals.  In contrast, EdgeBoxes only uses the information of edge, which is less effective in generating correct object proposals since it can be easily influenced by motion blur, low contrast, etc.  However, the EdgeBoxes algorithm is still better than the other three competing methods (i.e., CADM, MSTE and SelectiveSearch) since these three methods tend to generate correct object proposals for complete objects only, while a tracked target may be occluded and only a part of the target object can be seen in such a case.

The recall curves obtained by the five competing methods on the UAV20L dataset are shown in Fig~\ref{fig:ComparingRecallCurves}(c). As can be seen, the proposed TOPG method outperforms
 the other four competing methods, especially when only the top 50 ranked object proposals are used to calculate the recall. Table~\ref{tab:uav20l} shows the quantitative comparison results. The recall obtained by the proposed TOPG method is higher than that of the other four competing methods by about 18\%-60\% when only the top 50 ranked object proposals are used to calculate the recall. Generally, the performance ranking of the five competing methods is similar to that on the OTB-2013 dataset. However, the recall results obtained by the five competing methods on the UAV20L dataset are worse than those on the OTB-2013 dataset, especially for the case that more than 200 ranked object proposals are used. The main reason is that the UAV20L dataset contains many challenging video sequences, where the size of objects is very small.
%


Fig.~\ref{Fig:proposalDe} shows several qualitative results obtained by the proposed TOPG method on the OTB-2013 and UAV20L datasets. As can be seen, the proposed TOPG method can effectively hit the targets even when only the top 10 ranked object proposals are used, since TOPG can effectively re-rank object proposals by using multiple prior information of the tracked targets. Moreover, the generated object proposals by using TOPG inherently include better scale estimation for the targets.
\begin{figure}
\centering
   \includegraphics[width=0.90\linewidth]{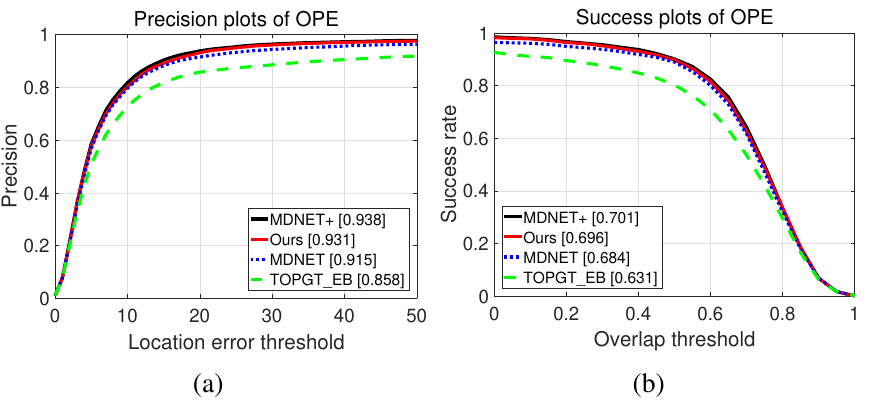}
   \caption{The precision and success plots obtained by several variants of the proposed tracker and the MDNet tracker on the OTB-2013 dataset.}
\label{Fig:ablation}
\end{figure}
\subsection{Evaluation of TOPG for Visual Tracking}
In this subsection, we demonstrate the performance of the proposed TOPG-based tracker (TOPGT) and compare it against seven state-of-the-art visual trackers (i.e., ECO~\cite{ECO17}, MDNet~\cite{MDNet2016}, SRDCF~\cite{martin2015}, KCF~\cite{Henriques2015}, SAMF~\cite{yangli2014}, KCFDP~\cite{kcfdp2015} and TLD~\cite{TLD2012}) on both the OTB-2013 dataset and the UAV20L dataset. The seven competing trackers are closely related to the proposed tracker, since all the competing trackers follow the framework of tracking-by-detection. Furthermore, KCFDP is based on object proposal generation. Among the competing trackers, MDNet is the winner of the VOT2015 challenge~\cite{VOT2015}, and ECO is a quite competitive tracker based on correlation filters and CNN features.


In the implementation of the proposed TOPGT, the size of the search window is set as $5$ times of that of the target window obtained from the previous frame. We use top 500 ranked object proposals generated by TOPG for visual tracking.  The threshold $\varphi$ and $\omega$ are respectively set to 0.7 and 0.5 to select positive and negative samples in the procedure of training the CNN classifier in the proposed TOPGT.

\begin{figure*}
\begin{center}
   \includegraphics[width=0.90\linewidth]{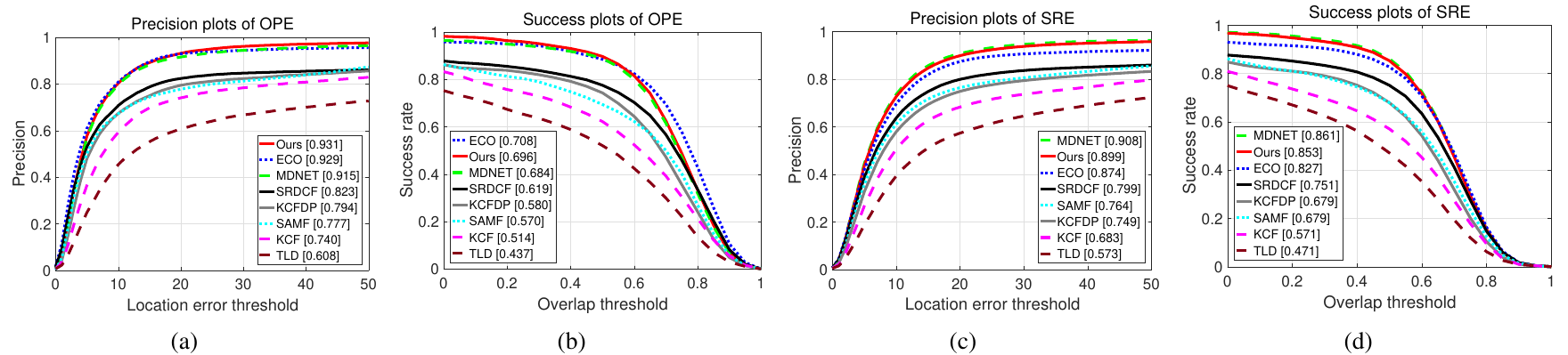}
\end{center}
   \caption{\label{Fig::PRECISION}The precision and success plots of the eight competing trackers on the OTB-2013 dataset with different evaluation criteria. (a)-(b) The OPE results obtained by the eight competing trackers. (c)-(d) The SRE results obtained by the eight competing trackers.}
\end{figure*}


\begin{figure*}
   \begin{center}
   \includegraphics[width=0.90\linewidth]{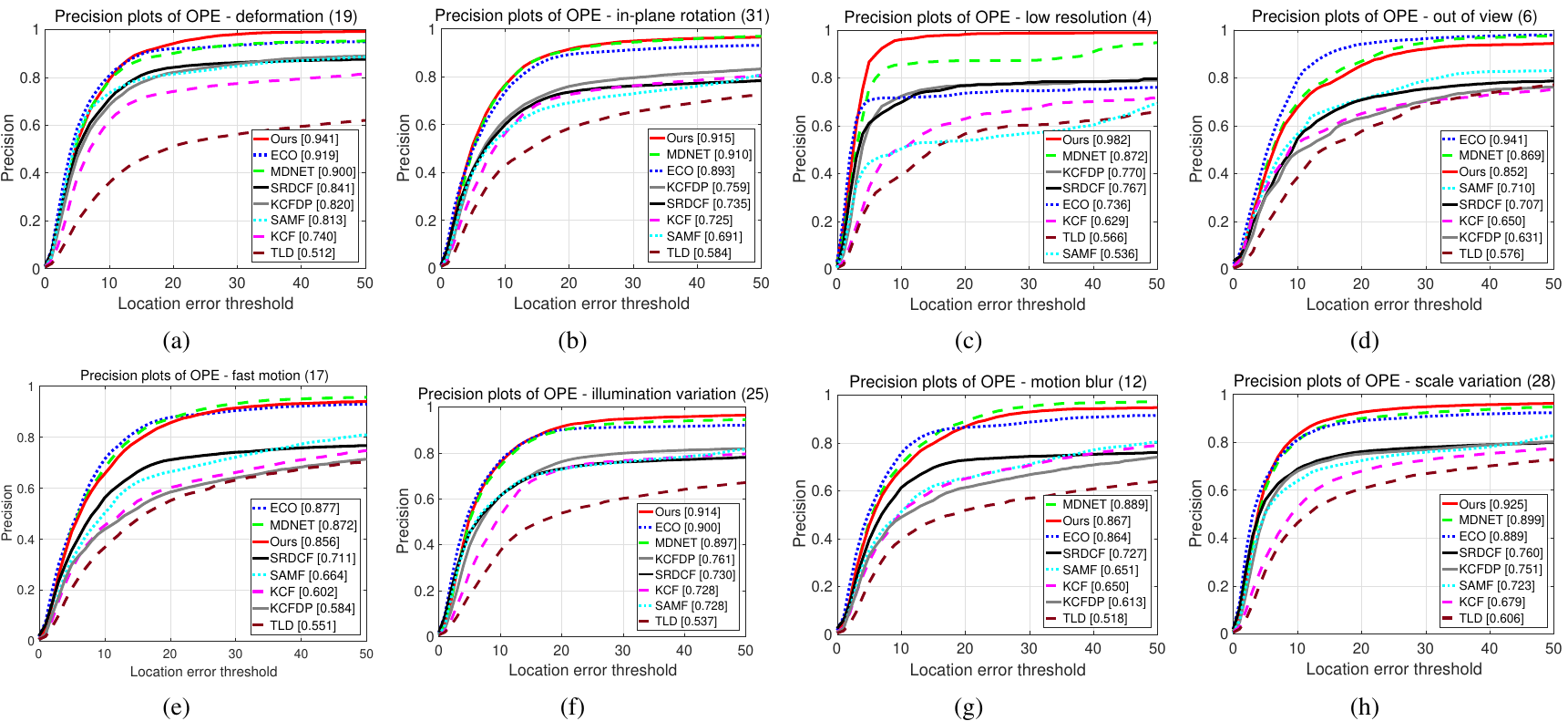}
\end{center}
   \vspace{-0.25cm}
   \caption{\label{Fig::ATTRIBUTE}The precision and success plots of the eight competing trackers on the eight attributes (i.e.,  (a) \emph{deformation},  (b) \emph{in-plane rotation}, (c) \emph{low resolution}, (d) \emph{out of view}, (e) \emph{fast motion}, (f) \emph{illumination variation}, (g) \emph{motion blur}, (h) \emph{scale variation}) of the video sequences in the OTB-2013 dataset with OPE.}
\end{figure*}
\subsubsection{Ablation Analysis}
Several components (such as the components of object proposal generation and the classifier) are involved in the proposed TOPGT.
In this subsection, we examine how these components affect the performance of the proposed TOPGT for visual tracking.

EdgeBoxes is used as a baseline object proposal generation algorithm. Thus, we also apply the Edgeboxes algorithm to the proposed TOPGT tracker and evaluate this variant as a comparison.  To do so, we replace the object proposals generated by TOPG with those generated by Edgeboxes, while the other components in the proposed TOPGT tracker are the same. For convenience, we call the above variant as TOPGT\_EB. Moreover, we apply the object proposals generated by the proposed TOPG method to MDNet as an effective sample selection component. In the test procedure of MDNet, hundreds of target candidates are generated from a 2D Gaussian distribution whose mean and variance are calculated by using the width and the height of the target obtained from the previous frame (please refer to~\cite{MDNet2016} for more details). However, the target candidates generated from a specified Gaussian distribution may not contain the tracked target when the target suffers from the challenges of occlusions, fast motion, deformation, etc. In contrast, the object proposals generated by the proposed TOPG method usually contain the tracked target. Thus, we can improve MDNet by collecting the object proposals generated by TOPG and the target candidates sampled from the Gaussian distribution as new target candidates, while the other parameter settings in MDNet are fixed. For convenience, we called the improved MDNet tracker as MDNet+.
Fig.~\ref{Fig:ablation}(a) shows the precision plots of the DP results obtained by TOPGT\_EB, MDNet, MDNet+ and the proposed tracker, respectively. As can be seen, TOPGT\_EB obtains the worst performance since the target recall obtained by using the Edgeboxes algorithm is low. In contrast, the proposed tracker obtains better performance than TOPGT\_EB. The reason is that the proposed tracker uses the proposed TOPG method instead of the Edgeboxes algorithm as an effective sample selection strategy. The object proposals generated by our TOPG method usually contain hard negative samples (e.g., similar objects), which are not easily selected by the other sample selection strategies. When the object proposals generated by TOPG are used to train the classifier in the proposed tracker, the trained classifier has more discriminative ability for similar objects than the classifier trained in TOPGT\_EB. In addition, the proposed TOPG method increases the target recall in the procedure of target detection, which reduces the probability of drift of the proposed tracker. Therefore, the proposed tracker obtains better performance than TOPGT\_EB.
The precision obtained by MDNet+ is higher than that of MDNet by about 2\%, which means TOPG can also be used to improve the MDNet tracker. Of course, the proposed TOPG method can be used by some other existing trackers as a key component to improve the performance of those trackers.  The success plots of OPE results obtained by TOPGT\_EB, MDNet, MDNet+ and the proposed tracker are shown in Fig.~\ref{Fig:ablation}(b). As can be seen, the proposed tracker and MDNet+, which is improved by using the proposed TOPG method, still obtain better performance.

\subsubsection{Compare with Several State-of-the-art Trackers}
\begin{figure*}
\begin{center}
\includegraphics[width=0.90\linewidth]{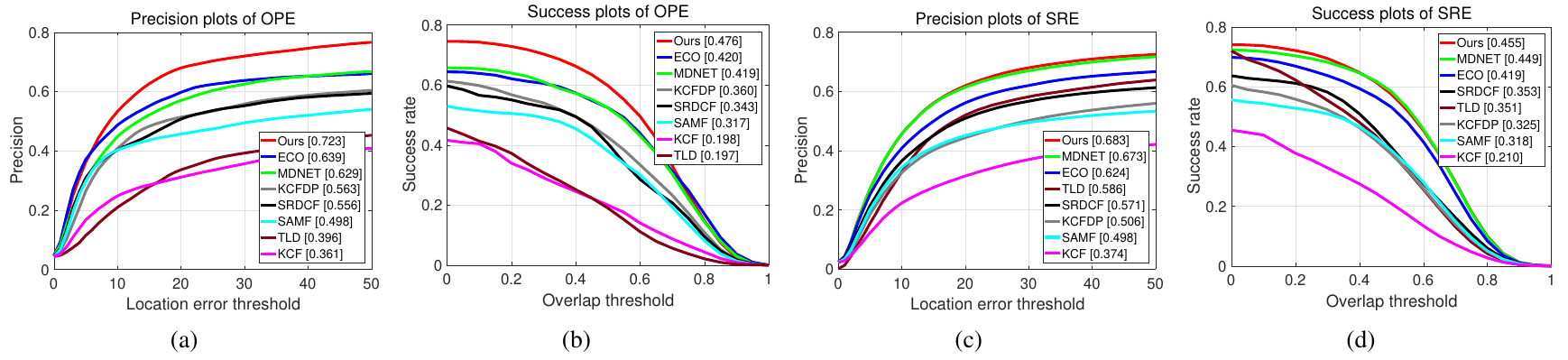}
\end{center}
   \vspace{-0.251cm}
   \caption{\label{Fig::UAVPRECISION}The precision and success plots of the eight competing trackers on the UAV20L dataset with different evaluation criteria. (a)-(b) The OPE results obtained by the eight competing trackers. (c)-(d) The SRE results obtained by the eight competing trackers.}
\end{figure*}
\begin{figure*}
\begin{center}
   \includegraphics[width=0.90\linewidth]{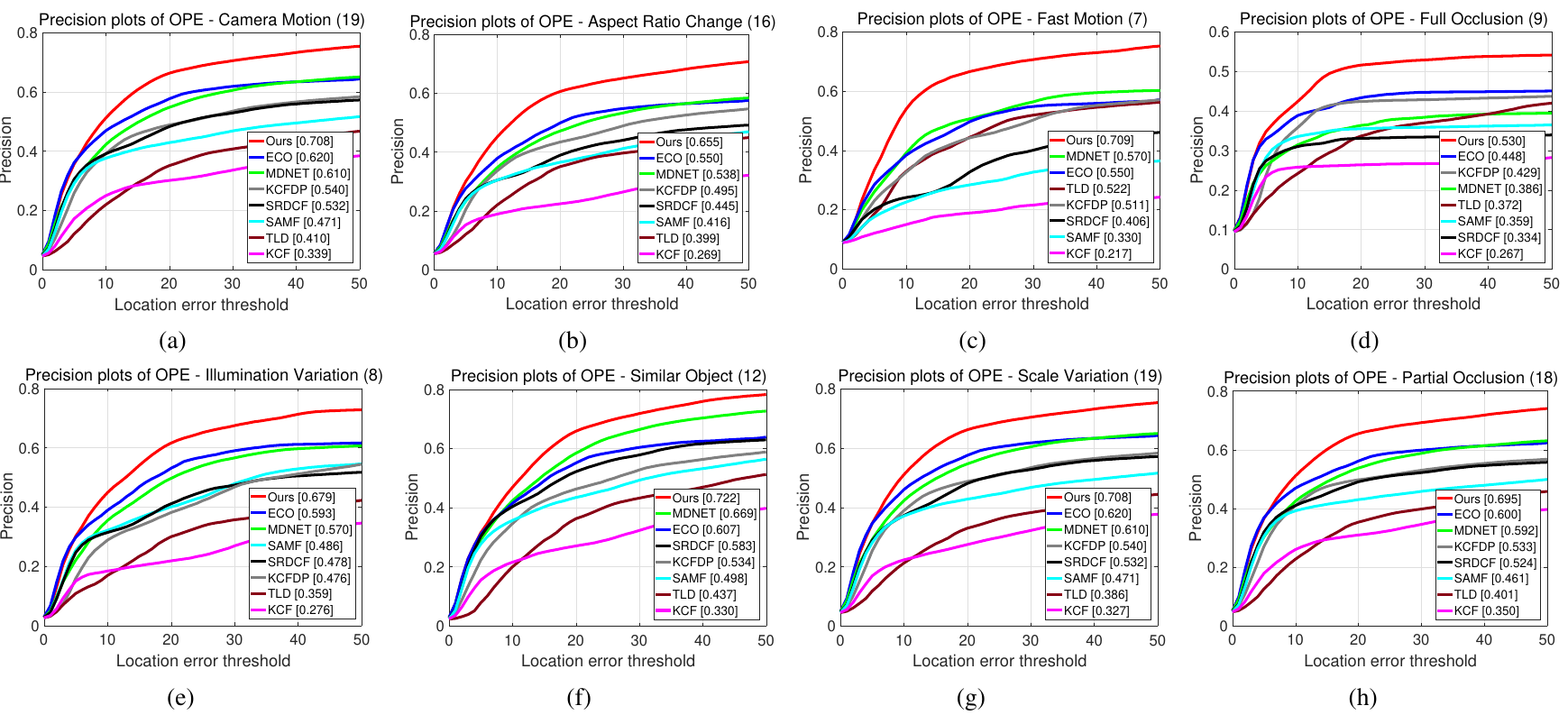}
\end{center}
   \vspace{-0.251cm}
   \caption{\label{Fig::UAVATTRIBUTE}The precision plots of the eight competing trackers on the video sequences with eight attributes (i.e.,  (a) \emph{camera motion}, (b) \emph{aspect ratio change}, (c) \emph{fast motion}, (d) \emph{full occlusion}, (e) \emph{illumination variation}, (f) \emph{similar object}, (g) \emph{scale variation}, (h) \emph{partial occlusion}) in the UAV20L dataset with OPE.}
\end{figure*}
Fig.~\ref{Fig::PRECISION} shows the precision and success curves  obtained by the eight competing trackers using the OPE and SRE evaluation criteria on the OTB-2013 dataset. As can be seen,  the proposed tracker outperforms the other competing six trackers (except for ECO) in terms of OPE, and outperforms the other six competing trackers (except for MDNet) in terms of SRE. As shown in Fig.~\ref{Fig::PRECISION}(a), the precision result obtained by the proposed TOPGT in OPE is $93.1\%$, which is better than those obtained by all other seven competing trackers. Moreover, the precision obtained by the proposed TOPGT is higher than that of the KCF tracker by 19.1\%.  Note that KCFDP uses the EdgeBoxes algorithm to generate object proposals for visual tracking. However, both the precision and the success rates obtained by the proposed TOPGT are significantly better than those of KCFDP, which means the proposed TOPG method is beneficial for visual tracking. The precision and success plots of the eight competing trackers in SRE are shown in Fig.~\ref{Fig::PRECISION}(c) and (d), respectively. As can be seen, the top two most accurate trackers are still MDNet and the proposed tracker.  Although the tracked targets are initialized by slightly shifting or scaling the ground truth with SRE, which means the proposed tracker may estimate incorrect probability distributions of the targets,  the proposed tracker still obtains competitive results. The reason is that the proposed TOPG method employed in TOPGT uses the trip-map scheme to alleviate the problem of mislabeling the pixels from a target and its background even when the tracked target is initialized with a small bias. MDNET obtains better performance than the proposed tracker in SRE since MDNet pre-trains a CNN classifier on a tracking dataset, which provides the CNN classifier with more discriminating power. One advantage of the proposed tracker is that it can be easily generalized to fit to other tracking datasets, since the CNN classifier in TOPGT is trained in an online manner.    In the remaining six competing trackers, the KCF tracker obtains higher DP values than TLD since the number of the training samples used in the KCF tracker is more than those used in TLD. However, the KCF tracker obtains lower DP values than the other improved correlation filter based trackers (i.e., SRDCF, KCFDP and SAMF), since these trackers have addressed some limits in the KCF tracker. For example, SRDCF learns a better classifier by selecting high-quality negative and positive samples using a spatial regularization strategy. The proposed TOPGT obtains excellent performance since it can effectively detect a tracked target. In addition, TOPGT also benefits from the good discriminating power of the CNN classifier.

Fig.~\ref{Fig::ATTRIBUTE} shows the precision and success results obtained by the eight competing trackers on the eight attributes (i.e., \emph{camera motion},  \emph{aspect ratio change},  \emph{fast motion}, \emph{full occlusion},  \emph{illumination variation}, \emph{similar object}, \emph{scale variation}, and \emph{partial occlusion}) of the video sequences in the OTB-2013 dataset. As can be seen, the proposed tracker is superior to most of the competing trackers (except for MDNet and ECO)  on these video sequences. The reason is that the proposed tracker locates a target by selecting the object proposals generated by TOPG, which is robust for target appearance variations and occlusion. More specifically, the response map, which is generated by using the color information, can efficiently suppress most of the noisy information from the background and it can highlight the target. Thus, high-quality object proposals can be generated from the  response map even when the target suffers from the problems of \emph{background clutter} or \emph{out-of-plane rotation}. Moreover, the response map is complementary to its original image for addressing the problem of weak edges caused by \emph{motion blur} and \emph{low contrast}. Thus, the proposed tracker obtains superior performance on the video sequences with \emph{motion blur}.

\begin{figure*}
\begin{center}
   \includegraphics[width=0.98\linewidth]{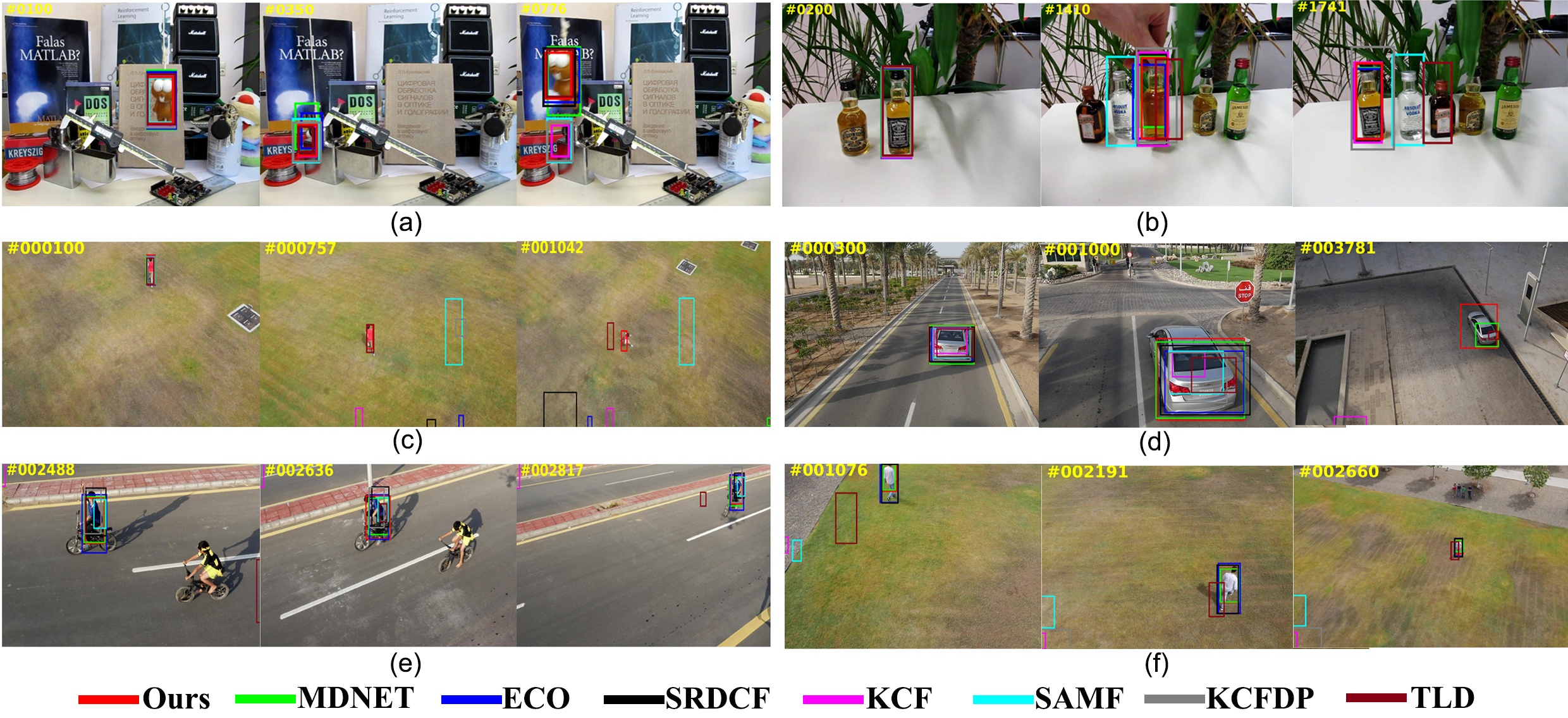}
\end{center}
   \vspace{-0.251cm}
   \caption{Several qualitative tracking results obtained by the eight competing trackers. (a)-(b) The obtained tracking results on the \emph{lemming} and \emph{liquor} video sequences in the OTB-2013 dataset, respectively. (c)-(f) The obtained tracking results on the  \emph{person7}, \emph{car6}, \emph{bike1} and \emph{person4} video sequences  in the UAV20L dataset, respectively.}
\label{Fig:demos}
\end{figure*}
To evaluate the performance of the eight competing trackers for long-term tracking, we test the eight competing trackers on the UAV20L dataset. Fig.~\ref{Fig::UAVPRECISION} shows the precision and success plots of the OPE results obtained by the eight competing trackers. As can be seen, the precision and success plots obtained by the proposed TOPGT in OPE are significantly higher than those obtained by the other competing trackers, especially higher than those obtained by MDNet about 9\% and 7\%, respectively. The main reason is that the proposed tracker utilizes the object proposals generated by the proposed TOPG method to train a robust CNN classifier and re-detect the target. Moreover, the object proposals generated by TOPG are robust for occlusions and deformation, which makes the proposed tracker effective for long-term tracking.  Among the competing trackers, ECO obtains better performance than the other correlation filter based methods (i.e., SRDCF, KCFDP, SAMF and KCF), since ECO effectively learns a convolution operator, which enables an elegant fusion of multi-resolution CNN feature maps for visual tracking. Fig.~\ref{Fig::UAVPRECISION} (c) and (d) show the precision and success plots of the SRE results obtained by the eight competing trackers on the UAV20L dataset. It is worthy of mentioning that the SRE evaluation criterion can not show the merits of the proposed tracker. However, the proposed TOPGT still obtains the best performance among the competing trackers.

We also evaluate the eight competing trackers on the video sequences with different attributes in the UAV20L dataset. Fig.~\ref{Fig::UAVATTRIBUTE} shows the precision plots of the OPE results obtained by the eight competing trackers on the video sequences with the eight attributes (i.e.,  \emph{camera motion},  \emph{aspect ratio change},  \emph{fast motion}, \emph{full occlusion},  \emph{illumination variation}, \emph{similar object}, \emph{scale variation} and \emph{partial occlusion}). As can be seen, the proposed tracker consistently outperforms the other seven competing trackers by a large margin. Especially, the DP obtained by the proposed TOPGT is higher than that obtained by MDNet for about 5\%-13\% on the video sequences with the eight attributes. The sample selection strategy in MDNet is less effective for several challenging situations, such as \emph{occlusions}, \emph{fast motion}, \emph{camera motion}, \emph{similar objects}, etc., since the tracked target may not be sampled by using a Gaussian distribution. Moreover, the classifier used in MDNet is not explicitly trained to distinguish the objects, which are similar to the tracked target. In contrast, the proposed tracker is quite effective in handling the above eight challenging situations, since the object proposals generated by the proposed TOPG are used to train the classifier in the TOPGT and re-detect the target. Fig.~\ref{Fig:demos} shows several qualitative tracking results obtained by the eight competing trackers. As can be seen, the proposed tracker is quite robust for different challenging situations.
\section{Conclusion}
In this paper, we have proposed a target-specific object proposal generation (TOPG) method, which combines the advantages of two important objectness cues (i.e., colors and edges) to deal with the weakness of each cue for generating object proposals for visual tracking. The integration of the two cues ensures that the proposed TOPG method achieves high recall of a specified target. We also give a  theoretical analysis to show that the proposed TOPG method can be easily integrated into other scoring-based object proposal generation methods to increase target recall. In addition, an effective object proposal ranking strategy is proposed, by which the proposed TOPG method significantly outperforms several other state-of-the-art object proposal generation methods in terms of recall and accuracy.  As experiments show, even that only 50 object proposals are used to evaluate, the proposed TOPG method still obtains higher recall than the other object proposal generation methods using 1,000 object proposals on two popular visual tracking datasets. Moreover, we apply the proposed TOPG method to the task of visual tracking, and propose a TOPG-based tracker (TOPGT) by treating the proposed TOPG method as an effective sample selection strategy in the procedures of both training a classifier and detecting the target. Since the object proposals generated by TOPG contain high-quality hard negative samples and positive samples, the proposed TOPGT is robust for different challenges in visual tracking. Experimental results show that the proposed TOPGT achieves surprisingly excellent tracking results and it outperforms most of the state-of-the-art trackers on long-term videos.


%

%
\appendices
\appendices
\numberwithin{equation}{section}
\section{The Analysis That TOPG Can Obtain Higher Recall Than the Scoring-based Object Proposal Generation Methods}
Assume that the number of object proposals is $N$, and each image contains only one target.  Let $g:\mathbf{x}\rightarrow \hat{\mathbf{y}} \subseteq \mathbb{R}^4$ denote an object proposal generation function, which can generate a set of object proposals $\{\hat{\mathbf{y}}_i\}_{i=1}^N$ from an image $\mathbf{x}$, where $i$ denotes the index of the object proposals. The optimal scoring-based object proposal generation methods are learned by maximizing the likelihood of correct object proposals as follows:
\begin{equation}
g^*=\mathop{\argmax}_{g\in \mathcal{G}}{\prod_{i=1}^N}{p_i(\mathop{max}_{\mathbf{y}_i\in g(\mathbf{x}_i)} \psi(\mathbf{y}_i,\hat{\mathbf{y}})\geq \varsigma \mid \mathbf{x},\mathbf{y})},
\label{Eq::optR} \end{equation}
where $p_i(\mathop{max}_{\mathbf{y}_i\in g(\mathbf{x}_i)} \psi(\mathbf{y}_i,\hat{\mathbf{y}})\geq \varsigma \mid \mathbf{x},\mathbf{y})$ (abbreviated as $p_i$) denotes the conditional probability that the $i$th object proposal generated by $g^*$ on the given image $\mathbf{x}$ hits the target; $\psi(\mathbf{y}_i,\hat{\mathbf{y}})$ measures the overlap between the $i$th object proposal $\hat{\mathbf{y}}_i$ and the ground-truth bounding box $\mathbf{y}$; $\varsigma$ denotes a threshold. Scoring-based object proposal generation methods usually  utilize scoring functions to give each object proposal a score according to a likelihood score how likely it contains an object. Thus, the $i$th object proposal usually hits a  target with a higher probability than the ($i+1$)th object proposal (i.e., $p_i > p_{i+1}$). Similarly to $p_i$, let $p_i^r$ denote the conditional probability of the $i$th object proposal generated by $g^*$ on the response map $\mathbf{x}^r$ hitting the target. Because that the response map $\mathbf{x}^r$ obtained by the proposed TOPG method can effectively suppress the noisy information and highlight the target, the search space for generating object proposals on the response map is significantly reduced. Thus,  the probability of the $i$th object proposal hitting the target in the response map is usually higher than that derived from the original image (i.e., $p_i^r > p_i$). To show that the top $N$ ranked object proposals generated by the proposed TOPG method can hit the target with a higher probability than that $N$ object proposals generated by $g^*$, the difference (denoted by $p^d$) between the probability of the object proposals generated by $g^*$ and that generated by TOPG is given as follows:
\begin{equation}
\begin{split}
p^d=p^g-p^t=
{\prod_{i=1}^N}{p_i}
-{\prod_{i=1}^{\frac{N}{2}}}{p_i}
*{\prod_{i=1}^{\frac{N}{2}}}{p_i^r} \\
=\prod_{i=1}^{\frac{N}{2}}{p_i}\bigg({\prod_{i=\frac{N}{2}+1}^N}{p_i}-{\prod_{i=1}^{\frac{N}{2}}}{p_i^r}\bigg)<0,
\end{split}
\label{Eq::probDiff} \end{equation}
where $p^g$ and $p^t$ respectively denote the probability of the correct detections obtained by the proposed TOPG method  and that of $g^*$. Note that the top $N$ ranked object proposals generated by proposed TOPG method include $\frac{N}{2}$ object proposals generated on an original image and $\frac{N}{2}$ object proposals generated on the response map.
The ratio $\tau$ between the minuend  and the subtractor in Eq.~(\ref{Eq::probDiff}) can be written as:
\begin{equation}
\begin{split}
\tau=\frac{{\prod_{i=\frac{N}{2}+1}^N}{p_i}}
{{\prod_{i=1}^{\frac{N}{2}}}{p_i^r}} <1.
\end{split}
\label{Eq::logProbDiff} \end{equation}
Since $p_{i+\frac{N}{2}}< p_i$ and $p_i <p_i^r$, $p_{i+\frac{N}{2}}< p_i^r$. Therefore, $\tau<1$, which means $p^g<p^t$. Thus, the proposed TOPG method can achieve higher recall with a higher probability than that of the existing scoring-based object proposal generation method, which is integrated into the proposed TOPG method.

%
%

\ifCLASSOPTIONcaptionsoff
  \newpage
\fi



%
%
%
\bibliographystyle{IEEEtran}
\bibliography{tsop}

%
\begin{IEEEbiography}[{\includegraphics[width=1in,height=1.25in,clip,keepaspectratio]{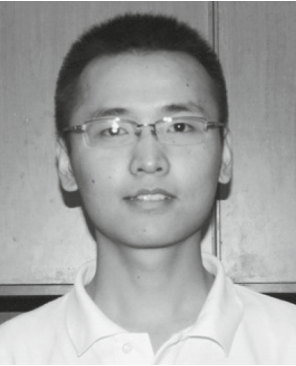}}]{Guanjun Guo}
is currently pursuing the Ph.D. degree with the Fujian Key Laboratory of Sensing and Computing for Smart City, and the School of Information Science and Engineering, Xiamen University, Xiamen, China.
His current research interests include computer vision, machine learning.
\end{IEEEbiography}
\begin{IEEEbiography}[{\includegraphics[width=1in,height=1.25in,clip,keepaspectratio]{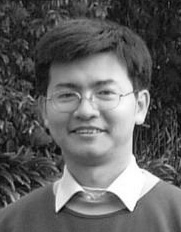}}]{Hanzi Wang}(SM'10)
is currently a Distinguished Professor of §Minjiang Scholars§ in Fujian province and a Founding Director of the Center for Pattern Analysis and Machine Intelligence (CPAMI) at XMU. He was an Adjunct Professor (2010-2012) and a Senior Research Fellow (2008-2010) at the University of Adelaide, Australia; an Assistant Research Scientist (2007-2008) and a Postdoctoral Fellow (2006-2007) at the Johns Hopkins University; and a Research Fellow at Monash University, Australia (2004-2006). He received his Ph.D degree in Computer Vision from Monash University where he was awarded the Douglas Lampard Electrical Engineering Research Prize and Medal for the best PhD thesis in the department. His research interests are concentrated on computer vision and pattern recognition including visual tracking, robust statistics, object detection, video segmentation, model fitting, etc. He has published more than 100 papers in major international journals and conferences including the IEEE T-PAMI, IJCV, ICCV, CVPR, ECCV, NIPS, MICCAI, etc.

He is a senior member of the IEEE. He was an Associate Editor for IEEE Transactions on Circuits and Systems for Video Technology (from 2010 to 2015) and a Guest Editor of Pattern Recognition Letters (September 2009). He was the General Chair for ICIMCS2014, Program Chair for CVRS2012, Publicity Chair for IEEE NAS2012, and Area Chair for ACCV2016, DICTA2010. He also serves on the program committee (PC) of ICCV, ECCV, CVPR, ACCV, PAKDD, ICIG, ADMA, CISP, etc, and he serves on the reviewer panel for more than 40 journals and conferences.
\end{IEEEbiography}
\begin{IEEEbiography}[{\includegraphics[width=1in,height=1.25in,clip,keepaspectratio]{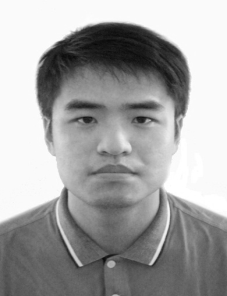}}]{Yan Yan}
is currently an associate professor in the School of Information Science and Technology at Xiamen University, China. He received the Ph.D. degree in Information and Communication Engineering from Tsinghua University, China, in 2009. He worked at Nokia Japan R\&D center as a research engineer (2009-2010) and Panasonic Singapore Lab as a project leader (2011). He has published around 40 papers in the international journals and conferences including the IEEE T-IP, T-Cyber, T-ITS, PR, KBS, Neurocomputing, ICCV, ECCV, ACM MM, ICPR, ICIP, etc. His research interests include computer vision and pattern recognition.
\end{IEEEbiography}

\begin{IEEEbiography}[{\includegraphics[width=1in,height=1.25in,clip,keepaspectratio]{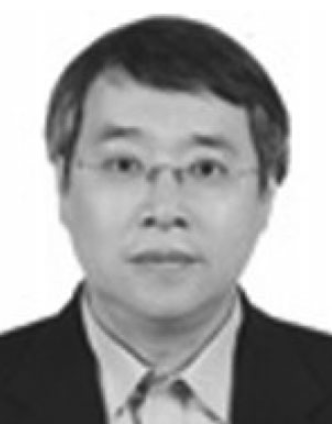}}]{Hong-Yuan Mark Liao} (F'13) received the PhD degree in electrical engineering from Northwestern
University, Evanston, IL, in 1990. In 1991, he joined the Institute of Information Science, Academia Sinica, Taipei, Taiwan, where he is currently a distinguished research fellow. He was in the fields of multimedia signal processing, image processing, computer vision, pattern recognition, video forensics, and multimedia protection for more than 25 years. He received the Young Investigators Award from Academia Sinica in 1998, the Distinguished Research Award from the National Science Council of Taiwan, in 2003, 2010, and 2013, respectively, the National Invention Award of Taiwan in 2004, the Distinguished Scholar Research Project Award from the National Science Council of Taiwan in 2008, and the Academia Sinica Investigator Award in 2010. His professional activities include the cochair of the 2004 International Conference on Multimedia and Exposition (ICME), the Technical cochair of the 2007 ICME, the General cochair of the 17th International Conference on Multimedia Modeling, the President of the Image Processing and Pattern Recognition Society of Taiwan (2006每2008), an Editorial Board member of the
IEEE Signal Processing Magazine (2010每2013), and an associate editor of the IEEE Transactions On Image Processing (2009每2013), the IEEE Transactions On Information Forensics And Security (2009每2012), and the IEEE Transactions On Multimedia (1998每2001). He also serves as the IEEE Signal Processing Society Region 10 Director (Asia-Pacific Region). He is a fellow of the IEEE.
\end{IEEEbiography}
\begin{IEEEbiography}[{\includegraphics[width=1in,height=1.25in,clip,keepaspectratio]{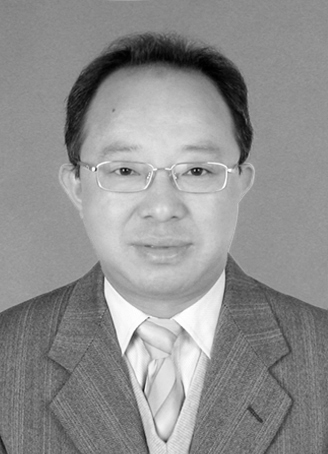}}]{Bo Li}
received the B.S. degree in computer science from Chongqing University in 1986, the M.S. degree in computer science from Xi＊an Jiaotong University in 1989, and the Ph.D. degree in computer science from Beihang University in 1993. Now he is a professor of Computer Science and Engineering at Beihang University, the Director of Beijing Key Laboratory of Digital Media, and has published over 100 conference and journal papers in diversified research fields including digital video and image compression, video analysis and understanding, remote sensing image fusion and embedded digital image processor.
\end{IEEEbiography}
\vfill




\end{document}